\documentclass[conference]{IEEEtran}
\usepackage{pifont}
\IEEEoverridecommandlockouts
\usepackage{cite}
\usepackage{amsmath,amssymb,amsfonts}
\usepackage{algorithmic}
\usepackage{array}
\usepackage{graphicx}
\usepackage{textcomp}
\usepackage{xcolor}
\usepackage{fancyhdr} 
\def\BibTeX{{\rm B\kern-.05em{\sc i\kern-.025em b}\kern-.08em
    T\kern-.1667em\lower.7ex\hbox{E}\kern-.125emX}}

\begin{document}

\title{Lightweight Multispectral Crop-Weed Segmentation for Precision Agriculture}

\author{\IEEEauthorblockN{Zeynep Galymzhankyzy}
\IEEEauthorblockA{\textit{Department of Math and Computer Science} \\
\textit{Lawrence Technological University} \\
Southfield, MI, USA \\
zgalymzha@ltu.edu}
\and
\IEEEauthorblockN{Eric Martinson}
\IEEEauthorblockA{\textit{Department of Math and Computer Science} \\
\textit{Lawrence Technological University} \\
Southfield, MI, USA \\
emartinso@ltu.edu}}

\maketitle

\fancypagestyle{withfooter}{
  \fancyhf{} 
  \renewcommand{\headrulewidth}{0pt} 
  \fancyfoot[C]{\footnotesize Accepted to the Novel Approaches for Precision Agriculture and Forestry with Autonomous Robots IEEE ICRA Workshop - 2025}
}

\thispagestyle{withfooter}
\pagestyle{withfooter}

\begin{abstract}
Efficient crop-weed segmentation is critical for site-specific weed control in precision agriculture. Conventional CNN-based methods struggle to generalize and rely on RGB imagery, limiting performance under complex field conditions. To address these challenges, we propose a lightweight transformer-CNN hybrid. It processes RGB, Near-Infrared (NIR), and Red-Edge (RE) bands using specialized encoders and dynamic modality integration. Evaluated on the WeedsGalore dataset~\cite{celikkan2025weedsgalore}, the model achieves a segmentation accuracy (mean IoU) of 78.88\%, outperforming RGB-only models by 15.8 percentage points. With only 8.7 million parameters, the model offers high accuracy, computational efficiency, and potential for real-time deployment on Unmanned Aerial Vehicles (UAVs) and edge devices, advancing precision weed management.

\end{abstract}


\section{Introduction}

Weed management is a critical challenge in precision agriculture, where accurate and timely identification of weeds enables site-specific control, reducing herbicide use and boosting crop yields. Recent advances in deep learning, particularly convolutional neural networks (CNNs), have improved crop–weed segmentation by analyzing field imagery. However, existing methods face significant limitations. Many rely solely on RGB imagery, which struggles to generalize across diverse vegetation types and lighting conditions, such as shadows or overcast skies. Additionally, static fusion strategies for combining multiple data types, like RGB and infrared, lack robustness to sensor noise or missing data, limiting their reliability in real-world settings. Transformer models, while effective at capturing complex patterns, are often too computationally intensive for deployment on resource-constrained Unmanned Aerial Vehicles (UAVs). Real-time performance, essential for practical UAV-based applications, remains underexplored in models using multispectral data, such as Near-Infrared (NIR) and Red-Edge (RE) bands.

To address these challenges, we propose a lightweight transformer–CNN hybrid model for efficient crop–weed segmentation. Our approach uses multispectral imagery (RGB, NIR, RE) through specialized encoders and a dynamic fusion mechanism that adapts to varying field conditions. With only 8.7 million parameters, the model is optimized for real-time inference on UAVs, offering high accuracy and computational efficiency.

\section{Related Work}

\subsection{Deep Learning Methods for Weed Segmentation}

Early approaches relied on handcrafted features and classifiers like Random Forests~\cite{haug2015crop}. While computationally efficient, these methods lacked robustness to field variability and required manual feature engineering.

Convolutional Neural Networks (CNNs) significantly improved segmentation accuracy. For WeedMap, Sa et al.~\cite{sa2018weedmap} used a modified SegNet, achieving strong performance for background and crop classes but lower accuracy for weeds. Celikkan et al.~\cite{celikkan2025weedsgalore} evaluated DeepLabv3+ and MaskFormer on WeedsGalore, achieving mean Intersection over Union (mIoU) scores above 82\% with multispectral inputs. However, CNNs often require deep architectures to model long-range dependencies, increasing computational demands and limiting efficiency on UAV platforms~\cite{wu2020robotic}.

Vision Transformers (ViTs) have recently gained traction for weed segmentation. Castellano et al.~\cite{castellano2023weed} proposed lightweight variants, Lawin and DoubleLawin, outperforming DeepLabv3 on WeedMap. Reedha et al.~\cite{reedha2022transformer} and Wang et al.~\cite{wang2023fine} showed that transformers excel at capturing global shape and context, crucial for distinguishing overlapping or visually similar vegetation. However, many ViTs remain computationally intensive, hindering real-time deployment.

\subsection{Multispectral Fusion Strategies}

Multispectral imagery provides spectral cues beyond RGB. NIR reflects plant health by capturing canopy structure, while RE detects chlorophyll variations, aiding crop–weed differentiation. Common fusion methods, such as channel stacking or concatenation, treat modalities equally, ignoring variations in quality or relevance.

WeedsGalore~\cite{celikkan2025weedsgalore} demonstrated that combining RGB, NIR, and RE improves segmentation over RGB-only models, particularly for rare weed classes. However, most fusion methods are static and cannot adapt to modality noise or dropouts. There is a need for adaptive fusion mechanisms that dynamically prioritize modalities based on their relevance or quality, enhancing robustness to environmental variability.

\subsection{UAVs and Real-Time Constraints}

UAVs are increasingly used to collect high-resolution, georeferenced multispectral data. Orthomosaic generation and pixel-level annotations enable large-scale field analysis, but real-time deployment remains challenging. CNNs and transformers often incur high memory and latency costs, limiting their use on resource-constrained UAV platforms.

Lightweight transformer--CNN hybrids offer a promising solution, balancing efficiency and performance with multispectral inputs, as demonstrated in tasks like image super-resolution~\cite{fang2022hnct}. Our proposed architecture addresses these challenges through modular spectral encoding, adaptive gated fusion, and efficient pyramid-based decoding, optimized for UAV-based precision agriculture.

\section{Multispectral and Multitemporal Data}

The WeedsGalore dataset~\cite{celikkan2025weedsgalore} is a publicly available benchmark for semantic and instance-level segmentation in maize fields, collected using an Unmanned Aerial Vehicle (UAV). Data was acquired over a \(1{,}840\;\mathrm{m}^2\) agricultural plot in Marquardt, Germany, with a DJI Phantom P4 Multispectral UAV. Four flights, conducted between May and June 2023, captured high-resolution imagery at a 5\,m altitude, achieving a ground sampling distance (GSD) of 2.5\,mm.

Each flight produced approximately 1,150 raw frames, from which 156 image tiles of \(600{\times}600\) pixels were manually cropped and annotated. The tiles are labeled across five semantic categories: maize, amaranth, barnyard grass, quickweed, and other weeds, representing common crops and weed species in Central European maize farming. The dataset is spatially split into 109 training (70\%), 23 validation (15\%), and 24 test (15\%) tiles, ensuring no spatial overlap between splits.

\begin{figure}[!t]
  \centering
  \includegraphics[width=0.9\columnwidth]{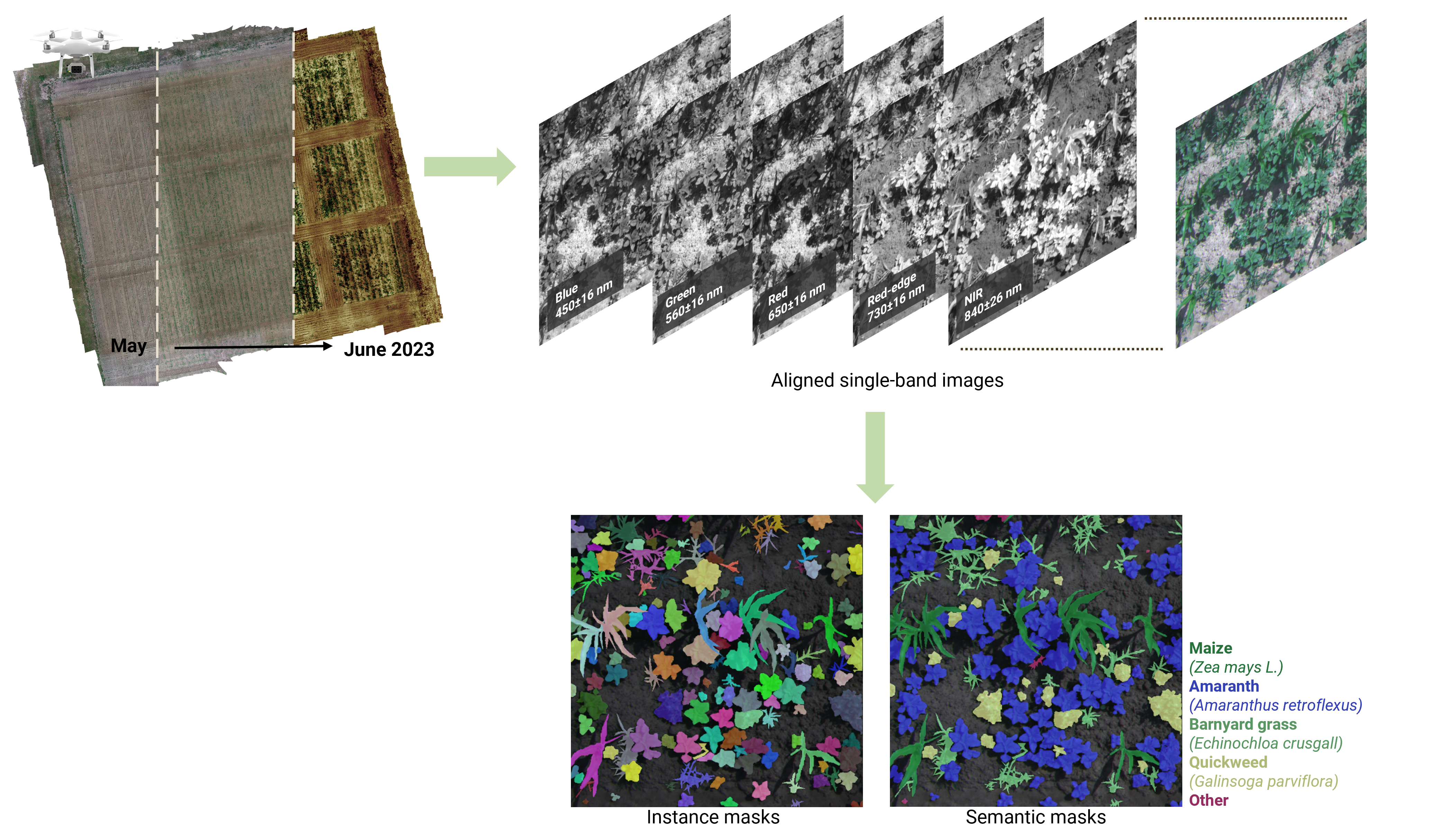}
  \caption{WeedsGalore dataset: acquisition and annotation workflow. Reprinted from~\cite{celikkan2025weedsgalore}.}
  \label{fig:wg_pipeline}
\end{figure}

WeedsGalore stands out due to several key features:
\begin{itemize}
    \item \textbf{Multispectral coverage}: RGB, Red-Edge (730\,nm), and Near-Infrared (NIR, 840\,nm) bands provide rich spectral information, enhancing crop–weed separability compared to RGB-only datasets.
    \item \textbf{Multitemporal sampling}: Four flights capture distinct plant growth stages, enabling models to learn robustly across temporal variations.
    \item \textbf{High annotation density}: Over 10,081 plant polygons, with an average of 78 instances per tile, offer a challenging benchmark for small-object segmentation, surpassing datasets like CWFID.
    \item \textbf{Public baselines}: Performance scores for DeepLabv3+ and MaskFormer, using both RGB and Multispectral Input (MSI), provide robust baselines for comparison.
\end{itemize}

For multimodal segmentation, each tile is processed into a 5-channel array (RGB, NIR, Red-Edge) and resized to \(600{\times}600\) pixels. Semantic masks are remapped into three categories: background, crop, and weed. On-the-fly augmentations, including random flips and rotations, are applied during training to improve generalization. This combination of spectral diversity, dense annotations, and temporal variation makes WeedsGalore ideal for evaluating dynamic modality weighting and Transformer-based segmentation models.

\begin{figure}[!t]
  \centering
  \includegraphics[width=0.9\columnwidth]{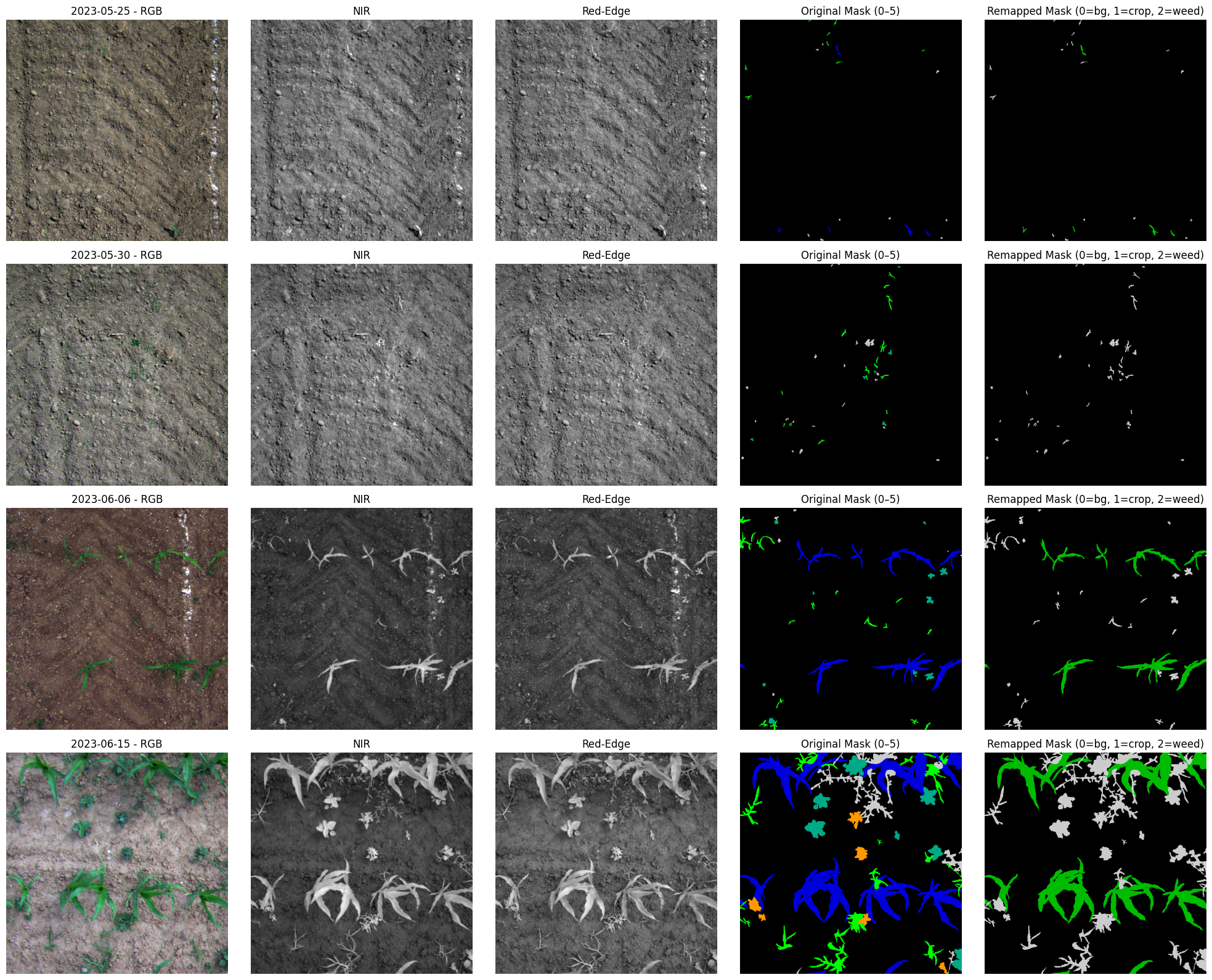}
  \caption{WeedsGalore dataset: example tile with RGB, NIR, and Red-Edge channels, alongside the remapped 3-class segmentation mask.}
  \label{fig:wg_sample}
\end{figure}

With its high-resolution 2.5\,mm GSD, multispectral richness, and multitemporal coverage, WeedsGalore is well-suited for developing site-specific weed management algorithms. By benchmarking on this dataset, we demonstrate that Transformer-driven multimodal fusion enhances segmentation accuracy while remaining efficient for real-time UAV inference.

\section{Proposed Methodology}
\begin{figure*}[!t]
    \centering
    \includegraphics[width=\textwidth]{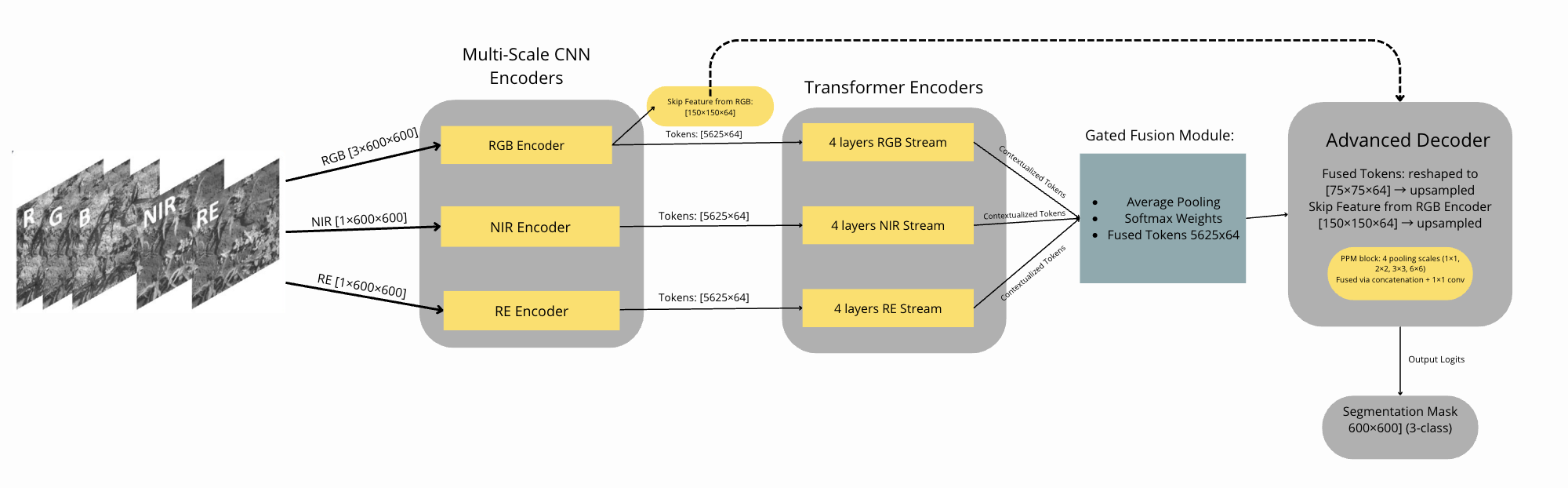}
    \caption{Architecture of the lightweight transformer--CNN hybrid, showing modality-specific encoding, Transformer refinement, gated fusion, and pyramid-based decoding.}
    \label{fig:full_pipeline}
\end{figure*}

\subsection{Model Architecture}
The proposed lightweight transformer–CNN hybrid is designed for multispectral crop–weed segmentation using Unmanned Aerial Vehicle (UAV) imagery. It processes five-channel inputs—RGB, Near-Infrared (NIR), and Red-Edge (RE)—through modality-specific convolutional encoders that extract low-level spatial features. Each modality stream is refined by dedicated Transformer blocks to capture long-range spatial dependencies.

A key component, the Gated Fusion Module, dynamically weighs each spectral modality using learned weights, prioritizing the most informative inputs under varying field conditions, such as sensor noise or lighting changes. The fused representation is processed by a Pyramid Pooling Module, which captures global context at multiple scales and integrates skip connections to refine fine-grained boundaries.

The final output is a high-resolution segmentation map distinguishing background, crop, and weed classes. With a compact footprint of 8.7 million parameters, the model is optimized for efficiency and suitable for real-time inference on embedded platforms, such as UAV-mounted NVIDIA Jetson devices. This balance of accuracy and computational efficiency makes it a strong candidate for site-specific weed management in precision agriculture. The full architecture is illustrated in Fig.~\ref{fig:full_pipeline}.


\subsection{Training and Evaluation}
To implement the model for crop–weed segmentation, we trained it on the WeedsGalore dataset~\cite{celikkan2025weedsgalore}, using 109 training and 23 validation tiles, each resized to \(600\times600\) pixels. A combination of Cross-Entropy Loss and Class-Balanced Focal Loss addressed weed class imbalance. The AdamW optimizer was employed with an initial learning rate of \(1\mathrm{e}{-4}\), cosine annealing, and a batch size of 8. Data augmentations, including random flips and rotations, were applied, and training ran for 100 epochs with mixed precision.

To evaluate performance, we assessed the model using mean Intersection-over-Union (mIoU), per-class IoU, and overall accuracy on a held-out test set of 24 images. All experiments maintained consistent preprocessing, augmentation, and class remapping into three categories: background, crop, and weed. The Multispectral Input (MSI) model, using RGB, NIR, and RE, achieved an mIoU of \textbf{78.88\%}, demonstrating robust segmentation of crop and weed regions. In contrast, the RGB-only variant yielded a lower mIoU of \textbf{63.08\%}, often misclassifying visually similar crop clusters as weeds, underscoring the importance of spectral diversity.

\begin{figure}[!t]
    \centering
    \includegraphics[width=0.9\columnwidth]{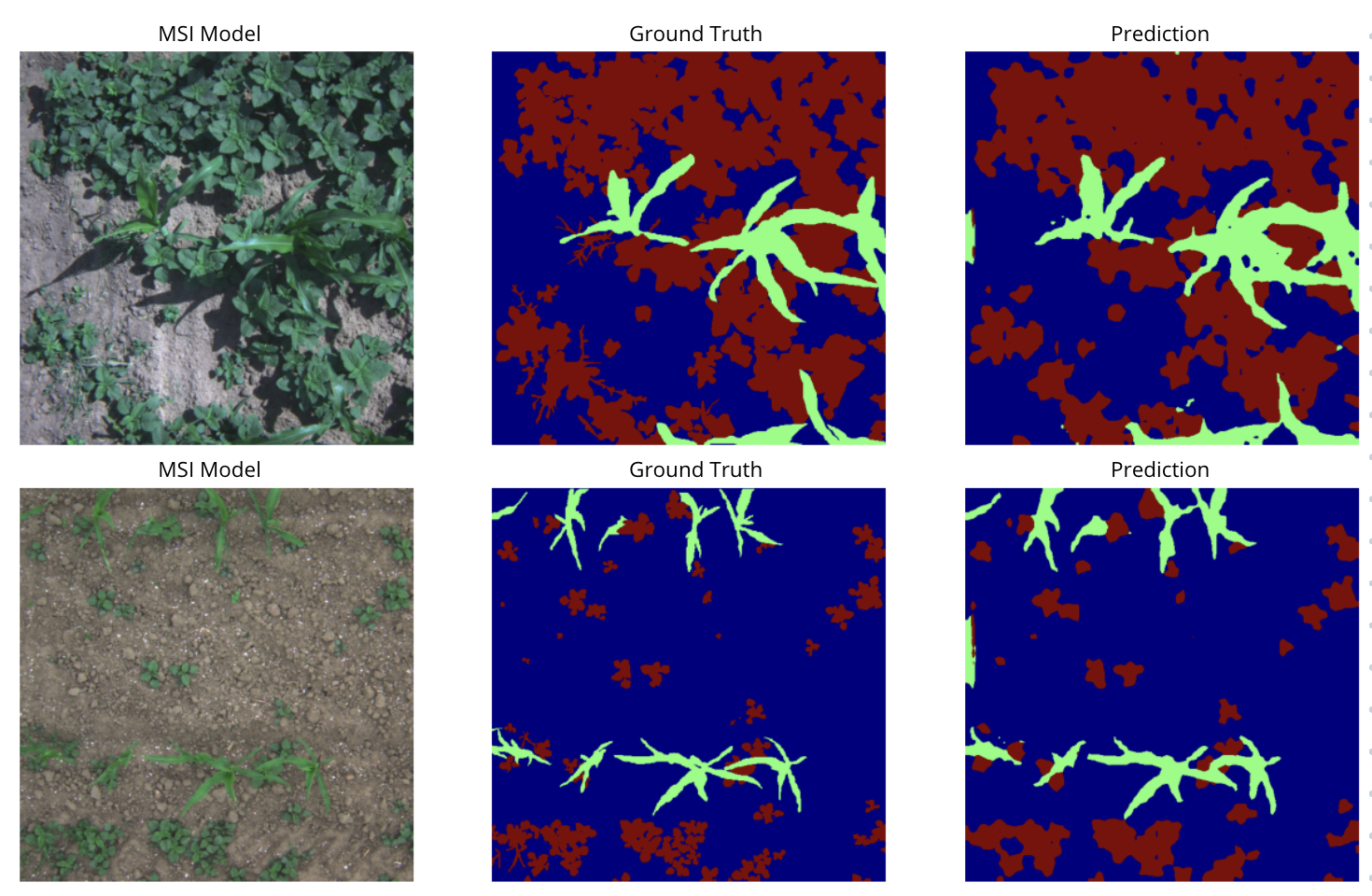}
    \caption{Segmentation by the MSI model: RGB input (left), ground truth (center), predicted mask (right). Note the accurate delineation of crop boundaries.}
    \label{fig:msi_result}
\end{figure}

\begin{figure}[!t]
    \centering
    \includegraphics[width=0.9\columnwidth]{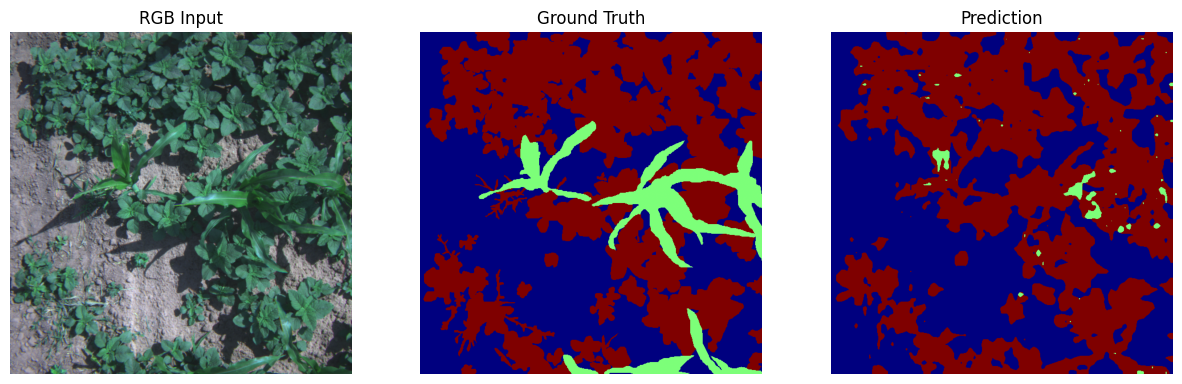}
    \caption{Segmentation by the RGB-only model: limited differentiation of dense weeds from crops.}
    \label{fig:rgb_only_result}
\end{figure}

Qualitative analysis further highlights these findings. As shown in Fig.~\ref{fig:msi_result}, the MSI model accurately captures fine-grained weed structures and maintains sharp boundaries. Conversely, the RGB-only model (Fig.~\ref{fig:rgb_only_result}) struggles in low-contrast scenes, reinforcing the value of multispectral features for reliable performance under real-world field conditions.

\subsection{Comparison with Baselines}
To contextualize the model’s performance, we compared it with two baselines from~\cite{celikkan2025weedsgalore}: DeepLabv3+ and MaskFormer. DeepLabv3+ achieves a higher mIoU of 82.90\% with multispectral input but relies on static early fusion and has a larger footprint of 41.2M parameters. Similarly, MaskFormer attains 79.55\% mIoU but is transformer-heavy, with 43.1M parameters, making it unsuitable for real-time UAV inference. In contrast, our model achieves 78.88\% mIoU with only 8.7M parameters, offering an efficient, modular alternative that dynamically fuses spectral cues. This trade-off between accuracy and efficiency positions the model as well-suited for embedded weed segmentation systems.

\begin{table}[!t]
\caption{Performance Comparison with Baselines (Multispectral Input)}
\label{tab:baseline_comparison}
\centering
\small
\begin{tabular}{|l|c|c|c|}
\hline
\textbf{Model} & \textbf{mIoU (\%)} & \textbf{Params (M)} & \textbf{Edge-Ready} \\
\hline
DeepLabv3+~\cite{celikkan2025weedsgalore} & 82.90 & 41.2 & No \\
MaskFormer~\cite{celikkan2025weedsgalore} & 79.55 & 43.1 & No \\
\textbf{Ours} & 78.88 & \textbf{8.7} & Yes \\
\hline
\end{tabular}
\end{table}

\section{Conclusion}

This work has proposed a lightweight transformer-CNN hybrid architecture to conduct multispectral crop–weed segmentation for precision agriculture. Using an adaptive gated fusion mechanism that dynamically weights spectral channels, it demonstrates enhanced robustness to environmental variations on the WeedsGalore dataset~\cite{celikkan2025weedsgalore}, achieving 78.88\% mIoU. This exceeds RGB-only models by 15.8 percentage points through the effective use of RGB, NIR, and RE bands for superior vegetation differentiation. With 8.7 million parameters, the architecture is also optimized for efficient inference to make it suitable for resource-constrained UAV platforms. Together, accuracy and efficiency make it a compelling solution for site-specific weed management, with potential for future edge deployment.

Proposed future work in this domain are twofold. First, the current training process is fully supervised training with dense, pixel-level annotations, making generalization to new crops or conditions laborious. These setup costs can be reduced by leveraging domain adaptation, few-shot learning, and/or self-supervised pre-training on unlabeled UAV data.  Additional training with modality dropout and redundancy-aware fusion should further enhance reliability under sensor failures.

The second goal is to deploy the architecture to edge devices as part of a precision agriculture solution.  Model pruning and hardware-aware Neural Architecture Search (NAS) should further reduce computational demands. Integrating the model into autonomous weed-control robots is also being investigated, combining segmentation with navigation and multi-sensor fusion like Light Detection and Ranging (LiDAR). Ultimately, long-term field trials under diverse conditions—varying weather, soil types, or farm scales—will be critical to enabling scalable, site-specific weed control.

\end{document}